%% file: main.tex
\documentclass[10pt,twocolumn,letterpaper]{article}

\usepackage{iccv}
\usepackage{times}
\usepackage{epsfig}
\usepackage{graphicx}
\usepackage{amsmath}
\usepackage{amssymb}
\usepackage{microtype}
\usepackage{enumitem}
\usepackage{dsfont}
\usepackage{bm}
\usepackage{booktabs}
\usepackage{amssymb}
\usepackage{pifont}
\newcommand{\cmark}{\ding{51}}%
\newcommand{\xmark}{\ding{55}}%
\usepackage[dvipsnames]{xcolor}
\usepackage{colortbl}
\usepackage{multirow}
\usepackage{multicol}
\usepackage{array}
\usepackage{soul}
\usepackage[accsupp]{axessibility}

\usepackage[pagebackref=true,breaklinks=true,letterpaper=true,colorlinks,bookmarks=false,citecolor=red]{hyperref}

\iccvfinalcopy 

\definecolor{Gray}{gray}{0.85}
\newcolumntype{a}{>{\columncolor{Gray}}c}
\def\iccvPaperID{5363} 

\ificcvfinal\fi

\begin{document}

\title{Prior-guided Source-free Domain Adaptation for Human Pose Estimation}

\author{Dripta S. Raychaudhuri$^{1,2 * }$ \ Calvin-Khang Ta$^{1 }$ \ Arindam Dutta$^{1 }$ \ Rohit Lal$^{1 }$ \ Amit K. Roy-Chowdhury$^{1 }$\\
$^{1}$University of California, Riverside $ \quad \quad ^{2}$AWS AI Labs \\
{\tt\small \{drayc001@, cta003@, adutt020@, rlal011@, amitrc@ece.\}ucr.edu} 
}

\maketitle
\ificcvfinal\thispagestyle{empty}\fi

\newcommand\blfootnote[1]{%
  \begingroup
  \renewcommand\thefootnote{}\footnote{#1}%
  \addtocounter{footnote}{-1}%
  \endgroup
}
\blfootnote{* Currently at AWS AI Labs. Work done while the author was at UCR.}

\input{sections/0_abstract.tex}
\input{sections/1_introduction.tex}

\input{sections/2_related.tex}

\input{sections/3_method.tex}
\input{sections/4_experiments.tex}

\input{sections/5_conclusion.tex}

{\small
\bibliographystyle{unsrt}
\bibliography{ref}
}

\newpage
\appendix

\input{sections/supp}

\end{document}

%% file: sections/0_abstract.tex
\begin{abstract}
Domain adaptation methods for 2D human pose estimation typically require continuous access to the source data during adaptation, which can be challenging due to privacy, memory, or computational constraints. To address this limitation, we focus on the task of source-free domain adaptation for pose estimation, where a source model must adapt to a new target domain using only unlabeled target data. Although recent advances have introduced source-free methods for classification tasks, extending them to the regression task of pose estimation is non-trivial. In this paper, we present \ul{P}ri\ul{o}r-guided \ul{S}elf-\ul{t}raining (\texttt{POST}), a pseudo-labeling approach that builds on the popular Mean Teacher framework to compensate for the distribution shift. \texttt{POST} leverages prediction-level and feature-level consistency between a student and teacher model against certain image transformations. In the absence of source data, \texttt{POST} utilizes a human pose prior that regularizes the adaptation process by directing the model to generate more accurate and anatomically plausible pose pseudo-labels. Despite being simple and intuitive, our framework can deliver significant performance gains compared to applying the source model directly to the target data, as demonstrated in our extensive experiments and ablation studies. In fact, our approach achieves comparable performance to recent state-of-the-art methods that use source data for adaptation.
\end{abstract}


%% file: sections/1_introduction.tex
\section{Introduction}

Human pose estimation is a fundamental task in computer vision that involves determining the precise locations of keypoints, such as joints, on a human body in an image or video~\cite{sun2019deep}. The growing need for pose estimation in various applications such as action recognition~\cite{yan2018spatial}, human-computer interaction~\cite{liu2022arhpe}, and video surveillance~\cite{li2020end} has driven the rapid development of highly accurate deep learning techniques. However, the challenge of obtaining large annotated datasets for training, compounded with the susceptibility to a performance decline in the face of data distribution shifts still poses limitations for current pose estimation models.

\begin{figure}[t]
    \centering
    \includegraphics[width=\columnwidth]{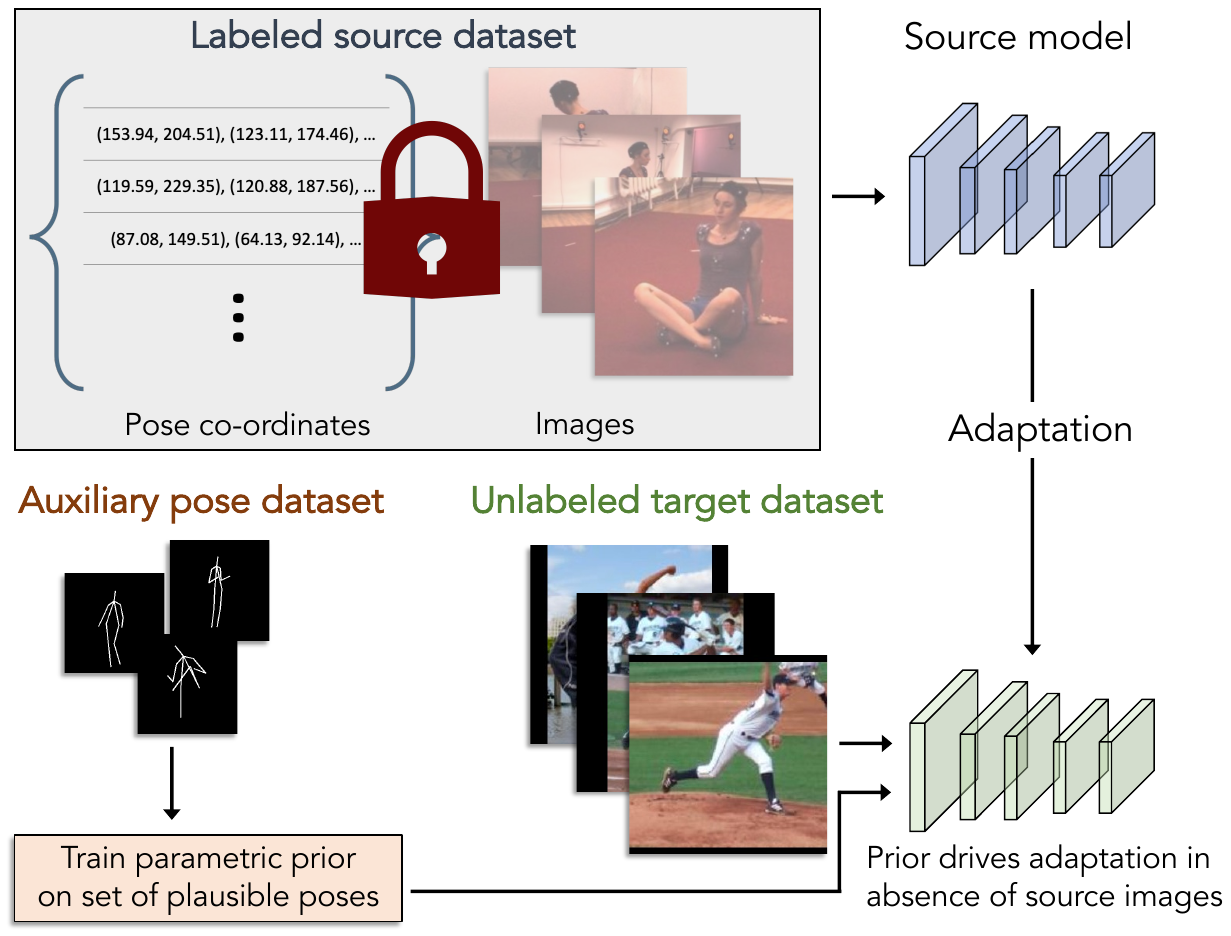}
    \caption{\textbf{Problem setup.} Existing UDA methods for pose estimation rely on a labeled source dataset while adapting to an unlabeled target dataset. However, privacy concerns surrounding the use of personally identifiable information in these labeled datasets, as well as the significant storage and computational requirements, can limit access to such data. Hence, our work focuses on source-free UDA of pose estimation models.}
    \label{fig:teaser}
\end{figure}

To overcome these limitations, recent studies have focused on unsupervised domain adaptation (UDA) of pose estimators~\cite{jiang2021regressive,kim2022unified}. UDA allows for transferring a pose estimation model trained on a source domain, where labeled data is available, to a target domain where labeled data is unavailable. Despite improved and robust pose estimation, the requirement of simultaneous access to both source and target domains during adaptation hinders real-world implementation. For instance, the labeled source data may not be accessible post-deployment due to privacy or proprietary issues. This is particularly relevant for human pose datasets, which contain \emph{personally identifiable information} (PII) \cite{schwartz2011pii}. 
Furthermore, adaptation using the entire source data might be infeasible due to both memory and computational constraints.
In light of these issues, we focus on \emph{source-free} UDA of human pose estimation models.

Concretely, our objective is to adapt a 2D human pose estimation model to a new target domain utilizing only its trained parameters and unlabeled target data. This presents a major challenge as the absence of source data for regularization can cause catastrophic forgetting. While recent advances have introduced methods to tackle this issue in classification tasks~\cite{liang2020we,ahmed2021unsupervised,paul2021unsupervised}, extending them to the regression task of pose estimation is non-trivial 
. To address this challenge, we introduce \ul{P}ri\ul{o}r-guided \ul{S}elf-\ul{t}raining (\texttt{POST}), a self-training regression framework that employs a human pose prior to effectively guide the adaptation process in the absence of source data. An overview of our problem setup is shown in Figure~\ref{fig:teaser}.

Our approach builds on the Mean Teacher~\cite{tarvainen2017mean} framework, which uses consistency in the prediction space of a student and teacher model to produce trustworthy pseudo-labels and learn from the unlabeled target domain. To achieve this, we create two augmented views of each target image, varying in scale, spatial context, and color statistics. Aligned pose predictions from both models in both views are then obtained, and consistency between the predictions across the different views is encouraged to facilitate \emph{prediction space adaptation}. However, our empirical results show that relying solely on consistency in the output space is insufficient when supervision from the source data is lacking. To address this, we also introduce \emph{feature space adaptation}, which aims to encourage consistency across features extracted from the two separate views. We adopt the Barlow Twins~\cite{zbontar2021barlow} approach to accomplish this. Specifically, we seek to make the cross-correlation matrix calculated from a pair of feature embeddings as close to the identity matrix as possible. 

In addition to the adaptation across both outputs and features, we employ a human pose prior that models the full manifold of plausible poses in some high-dimensional pose space to refine possible noisy pseudo-labels that may arise during self-training. The plausible poses are represented as points on the manifold surface, with zero distance from it, while non-plausible poses are located outside the surface, with a non-zero distance from it. This manifold is learned using a high-dimensional neural field, similar to Pose-NDF~\cite{tiwari22posendf}. The pose prior acts as a strong regularizer, directing the model to generate more accurate pose pseudo-labels on the target data and leading to improved adaptation. The learning of this prior requires an auxiliary dataset of plausible human poses, but this does not compromise the privacy aspect of our framework as the prior does not make use of RGB images. In addition, it is worth noting that the prior can be trained offline, separately from the adaptation process. This not only saves computational resources but also reduces the amount of storage required. Compared to storing entire images, it is much more efficient to store pose coordinates, which requires approximately $3000\times$ less memory.

\noindent \textbf{Main contributions.} To summarize, our primary contributions are as follows: 
\begin{itemize}[leftmargin=*,topsep=0pt]
\setlength\itemsep{-3pt}
    \item We address the problem of adapting a human pose estimation model to a target domain consisting of unlabeled data, without access to the original source dataset. This ameliorates the privacy concern associated with the current domain adaptive pose estimation methods.
    \item We introduce \ul{P}ri\ul{o}r-guided \ul{S}elf-\ul{t}raining (\texttt{POST}), a simple source-free unsupervised adaptation algorithm. \texttt{POST} leverages both prediction-level and feature-level consistency, in addition to a human pose prior, to drive self-training for improved adaptation to the target domain. 
    \item We evaluate our method qualitatively and quantitatively on three challenging domain adaptive scenarios, demonstrating comparable performance to existing UDA methods that have access to the source data.
\end{itemize}

%% file: sections/2_related.tex
\section{Related Works}
\noindent \textbf{Pose Estimation.}
2D human pose estimation aims to locate human anatomical keypoints, such as the elbow and knee. Prior works can be categorized into two primary frameworks: the top-down framework and the bottom-up framework. Top-down methods~\cite{fang2017rmpe,he2017mask,wei2016convolutional,newell2016stacked,xiao2018simple,chen2018cascaded,sun2019deep} first detect each person from the image and then perform single-person pose estimation on each bounding box independently. On the other hand, bottom-up methods~\cite{pishchulin2016deepcut,insafutdinov2016deepercut,cao2019openpose,kreiss2019pifpaf,cheng2020higherhrnet,geng2021bottom,jin2020differentiable} predict keypoints of each person directly in an end-to-end manner. Typical bottom-up methods consist of two steps: predicting keypoint heatmaps and grouping the detected keypoints into separate poses. In this work, we focus on the bottom-up framework for efficiency purposes and adopt the Simple Baseline~\cite{xiao2018simple} architecture following~\cite{kim2022unified} to ensure fair comparisons with prior domain adaptation algorithms. 
\vskip 6pt
\noindent \textbf{Unsupervised Domain Adaptation.}
UDA methods have been extensively applied to a broad range of computer vision tasks, including image classification~\cite{tzeng2017adversarial}, semantic segmentation~\cite{tsai2018learning}, object detection~\cite{hsu2020progressive}, and reinforcement learning~\cite{raychaudhuri2021cross} to tackle the issue of data distribution shift. Most approaches aim to align the source and target data distributions through techniques such as maximum mean  discrepancy~\cite{long2015learning} and adversarial learning~\cite{ganin2016domain,tzeng2017adversarial}. Another line of research utilizes image translation methods to perform adaptation by transforming the source images into the target domain~\cite{Hoffman_ICML_2018,Luan_CVPR_2019}. More recently, there has been a surge of interest in adaptation using only a pre-trained source model due to privacy and memory storage concerns related to the source data. These include techniques such as information maximization~\cite{liang2020we,ahmed2021unsupervised}, pseudo-labeling~\cite{yeh2021sofa,kumar2023conmix} and self-supervision~\cite{xia2021adaptive}. Compared to other tasks, domain adaptation for regression tasks, such as pose estimation, remains relatively unexplored. 
\vskip 6pt
\noindent \textbf{Domain Adaptive Pose Estimation.}
UDA methods for pose estimation have explored various techniques for overcoming the domain gap, including adversarial feature alignment and pseudo-labeling. RegDA~\cite{jiang2021regressive} estimates the domain discrepancy by evaluating false predictions on the target domain and minimizes it. Mu \emph{et al.}~\cite{mu2020learning} proposed consistency regularization with respect to transformations and temporal consistency learning within a video. Li \emph{et al.}~\cite{li2021synthetic} proposed a refinement module and a self-feedback loop to obtain reliable pseudo-labels. Recently, Kim \emph{et al.}~\cite{kim2022unified} introduced a unified framework for both human and animal keypoint detection, which aligns representations using input-level and output-level cues. Typically, these methods require access to the source data, which may raise data privacy, memory and computation concerns. In contrast, our method addresses the domain adaptation problem in a source-free setting.

%% file: sections/3_method.tex
\begin{figure*}[t]
    \centering
    \includegraphics[width=0.85\textwidth]{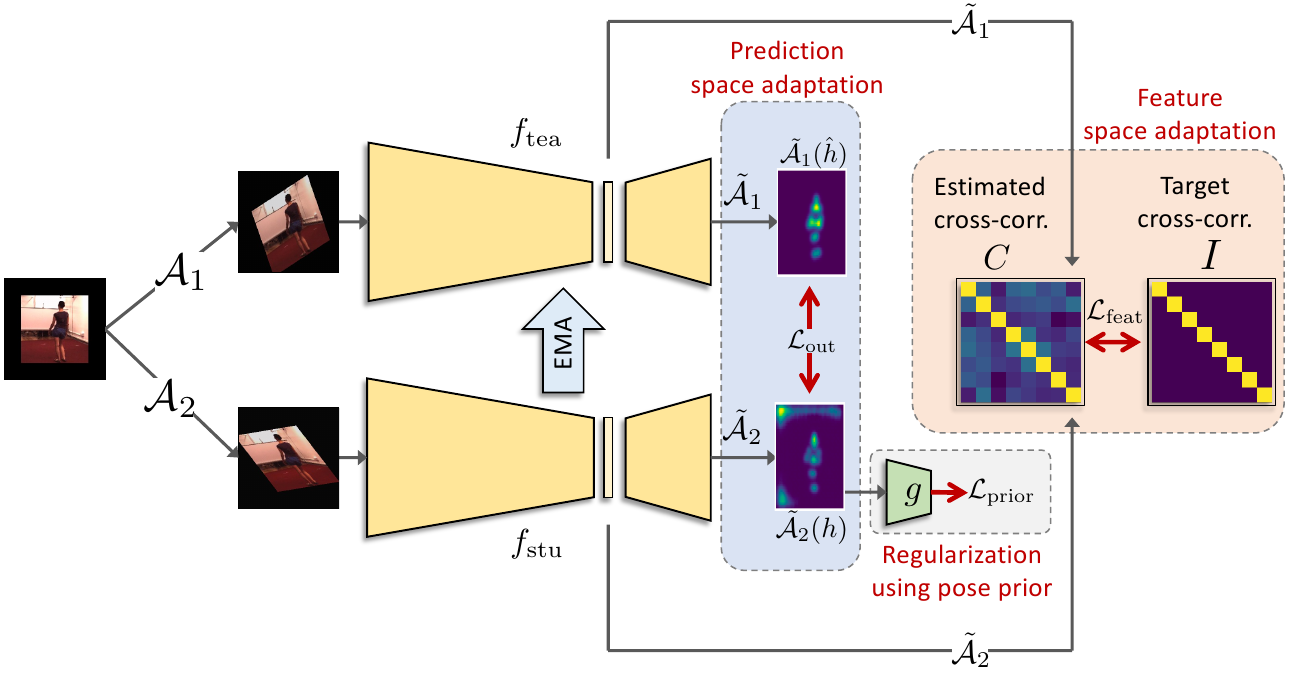}
    \caption{\textbf{Framework overview.} Our approach builds on the Mean Teacher framework and performs adaptation both in the pose prediction space using $\mathcal{L}_\textrm{out}$, and the feature space using $\mathcal{L}_\textrm{feat}$. This is supplemented by a human pose prior $g$ that scores the predicted pseudo-labels in terms of plausibility. These scores are used to regularize the adaptation process in the absence of labeled source data via $\mathcal{L}_\textrm{prior}$. The student model $f_\textrm{stu}$ is trained by the combination of the three losses, while the teacher model $f_\textrm{tea}$ is updated with the exponential moving average (EMA) of the weights of the student model. }
    \label{fig:framework}
\end{figure*}

\section{Prior-guided Self-training}
Our work investigates source-free domain adaptation for 2D human pose estimation. In the pose estimation task, given an input image $\bm{x}\in\mathbb{R}^{H\times W\times 3}$, the goal is to predict the corresponding $y\in\mathbb{R}^{K\times 2}$ representing the 2D coordinates of $K$ keypoints using a pose regression model $f$. In this paper, we assume access to a pre-trained model, denoted by $f_\mathcal{S}$, as well as $N$ unlabeled images $\mathcal{D}=\{\bm{x}_i\}_{i=1}^N$ from a target domain $\mathcal{T}$. Our goal is to adapt the source model to the target such that it performs better on images drawn from the target distribution than when directly using the source model on the target images.
\vskip 4pt
\noindent \textbf{Overview.} 
In the absence of source data, we propose to use self-training to adapt the source pose estimation model to the target domain. However, self-training methods are prone to error accumulation, particularly when labeled data is absent to act as regularization. Hence, we introduce \texttt{POST}, an enhanced self-training strategy that employs three essential ideas to prevent such errors:
\begin{enumerate}[leftmargin=*,topsep=0pt]
\setlength\itemsep{-2pt}
    \item A weight-averaged teacher model is used to generate the pseudo-labels for self-training. This ensures better retention of the source knowledge within the teacher model by reducing the effect of updating the weights via noisy pseudo-labels (Section~\ref{sec:output_adapt}).
    \item In addition to adaptation over the output space via pseudo-labels, the model is adapted in the feature space as well. For each target image, two aligned predictive views are generated via data augmentation, and consistency across features extracted from the two separate views is encouraged via a contrastive learning strategy (Section~\ref{sec:feat_adapt}).
    \item A human pose prior is used to regularize the adaptation by directing the model to generate more accurate and anatomically plausible pose pseudo-labels (Section~\ref{sec:prior}).
\end{enumerate}
An overview of our framework is presented in Figure~\ref{fig:framework}.

\subsection{Self-training via Mean Teacher} \label{sec:self_training}
Motivated by research suggesting that weight-averaged models over training steps tend to perform better than the final model~\cite{kim2022unified,li2021synthetic}, we utilize the Mean Teacher framework~\cite{tarvainen2017mean} to generate pseudo-labels for self-training. The framework involves creating two identical models, a teacher model $f_{\textrm{tea}}$ and a student model $f_{\textrm{stu}}$, both of which are initialized with the pre-trained network $f_\mathcal{S}$ at time step $t=0$. At each subsequent time step $t$, the student model parameters $\theta$ are updated by backpropagating the supervisory signals provided by the teacher model. The parameters of the teacher model $\Tilde{\theta}$ are updated via an exponential moving average (EMA) of the student model parameters:
\begin{equation} \label{eq:ema}
    \Tilde{\theta}_t = \alpha \Tilde{\theta}_{t-1} + (1-\alpha) \theta_t,
\end{equation}
where $\alpha$ denotes the smoothing coefficient which is set to $0.999$ by default. The EMA update prevents the teacher model from overfitting to noisy pseudo-labels during the initial rounds of self-training, thereby, preserving the source knowledge. This is especially advantageous in our scenario where source data is unavailable to regularize the adaptation. 

In the following sections, we demonstrate how to adapt $f_{\textrm{stu}}$ using supervisory signals from $f_{\textrm{tea}}$ on both the feature space and the pose prediction space. 

\subsubsection{Prediction Space Adaptation} \label{sec:output_adapt}
At each time step $t$, we apply two different data augmentations, $\mathcal{A}_1$ and $\mathcal{A}_2$, to a target image $\bm{x}$ to generate two views. We then obtain the keypoint heatmap corresponding to the first transformed image $\Tilde{h}^t=f_{\textrm{tea}}^t(\mathcal{A}_1(\bm{x}))$. Here $\Tilde{h}^t \in \mathbb{R}^{K\times H'\times W'}$ denotes the spatial likelihood of the $K$ different keypoints on each channel. The pseudo-label from the teacher model is generated by obtaining the coordinates which produce the maximum activations $\hat{y}^t = \textrm{argmax}_{u}\Tilde{h}^t_{[:,u]}$. 

To reduce the influence of erroneous pseudo-labels in our training process, we utilize a confidence threshold to discard potentially unreliable labels. Specifically, we only retain the keypoint activations with the top $p\%$ maximum values among all activations and discard the rest. We set the threshold $\tau$ accordingly to reflect this, thereby ensuring that only the most confident labels are used for training.

Following prior work on supervised pose estimation, we first convert the pseudo-labels to normalized Gaussian heatmaps~\cite{tompson2014joint} and then use the mean squared error (MSE) loss to update the student model over batches of target images sampled from $\mathcal{D}$:
\begin{equation} \label{eq:out}
    \displaystyle
    \mathcal{L}_{\textrm{out}} = \frac{1}{|\mathcal{B}|}\sum_{\mathbf{x}\in\mathcal{B}}\sum_{k=1}^K \mathds{1}\left(\Hat{h}^t_k\geq\tau\right)\Vert \Tilde{\mathcal{A}_1}(\Hat{h}^t_k) - \Tilde{\mathcal{A}}_2(h^t_k) \Vert_2.
\end{equation}
We denote the inverse functions of the chosen augmentations $\mathcal{A}_1$ and $\mathcal{A}_2$ as $\Tilde{\mathcal{A}_1}$ and $\Tilde{\mathcal{A}_2}$, respectively. The heatmap predicted by the student model for the $k$-th keypoint is represented by $h^t_k = f^t_{\textrm{stu}}(\mathcal{A}_2(\bm{x}))^k$, while the heatmap generated from the pseudo-labels predicted by the teacher model is represented by $\Hat{h}^t_k=L(\hat{y}^t)^k$. Here, $L(\cdot)$ represents the heatmap generating function and $\mathcal{B}$ denotes a batch of target images.

\subsubsection{Feature Apace Adaptation} \label{sec:feat_adapt}
Pose estimation models primarily rely on a high-to-low-resolution feature encoder to generate low-resolution representations, which are then used to recover high-resolution pose heatmaps~\cite{xiao2018simple}. Since the upsampling process is inherently noisy, providing intermediate supervision to explicitly adapt the features, in addition to adapting the output pose keypoints as shown in the previous section, can be beneficial~\cite{newell2016stacked}. While previous work has demonstrated the benefits of joint adaptation across the output and feature space via adversarial learning in tasks such as semantic segmentation~\cite{tran2019gotta}, this has been exclusively focused on scenarios where source data is available. Here, we propose an alternative way to accomplish feature space adaptation in the absence of source data via contrastive learning.

We begin by creating two different views of each target image $\mathbf{x}$ using a pair of sampled augmentations $\mathcal{A}_1$ and $\mathcal{A}_2$, as previously shown. Next, considering the pose estimation model as a composition of a feature encoder and an output regressor, \emph{i.e.}, $f=\textrm{Dec}\circ\textrm{Enc}$, we extract the augmentation reversed feature maps from the teacher model $\Tilde{z}=\Tilde{\mathcal{A}}_1(\textrm{Enc}_{\textrm{tea}}(\mathcal{A}_1(\bm{x})))$ and the student model $z=\Tilde{\mathcal{A}}_2(\textrm{Enc}_{\textrm{stu}}(\mathcal{A}_2(\bm{x})))$. We extract pairs of features for every image in a batch, pool them along the spatial dimensions, and then normalize them along the batch dimension to ensure that each covariate has a mean of 0 over the batch. For simplicity, we overload $z,\Tilde{z}$ to represent the normalized features, and drop the time index $t$. 

We utilize feature-level consistency between the different views in order to accomplish feature space adaptation. This is achieved via a contrastive learning strategy which encourages the cross-correlation matrix between the outputs of the two networks to be as close to the identity matrix as possible~\cite{zbontar2021barlow},
\begin{equation} \label{eq:feat}
\mathcal{L}_{\textrm{feat}} = \sum_i  (1-C_{ii})^2  + \gamma \sum_{i}\sum_{j \neq i} {C_{ij}}^2.
\end{equation}
We define $\gamma$ as a positive constant that balances the importance of the first and second terms of the loss function. $C$ represents the cross-correlation matrix computed between the outputs of the student and teacher networks along the batch dimension:
\begin{equation} \label{eq:crosscorr}
C_{ij} = \frac{
\sum_{b=1}^{|\mathcal{B}|} \Tilde{z}_{b,i}z_{b,j}}
{\sqrt{\sum_{b=1}^{|\mathcal{B}|} {(\Tilde{z}_{b,i})}^2} \sqrt{\sum_{b=1}^{|\mathcal{B}|} {(z_{b,j})}^2}}.
\end{equation}
Here, $b$ represents the batch index, while $i$ and $j$ index the feature dimensions of the networks' outputs. We set $\gamma=5e-3$ following~\cite{zbontar2021barlow}. 

The first term of $\mathcal{L}_{\textrm{feat}}$ encourages the consistency of pose features within the same image by equating the diagonal elements of the cross-correlation matrix to 1, effectively reversing the effects of augmentations. In contrast, the second term aims to decorrelate the different feature dimensions of the embedding by forcing the off-diagonal elements of the cross-correlation matrix to 0.

\subsection{Regularization via Pose Prior} \label{sec:prior}
Enhancing the performance of the pose estimation model on the target domain through joint adaptation over the output and feature space is undoubtedly valuable. However, this approach has its limitations, as it relies solely on general domain adaptation principles and overlooks the rich structural priors associated with human poses. To address this issue, we propose incorporating a parametric human pose prior to better adapt the pose estimation model to the target domain.

\subsubsection{2D Human Pose Prior}
Building upon the work of~\cite{tiwari22posendf}, we propose a human pose prior modeled as a manifold consisting of plausible 2D poses. To represent the 2D poses while ignoring aspects such as size and scale, we use a set of 2D orientation vectors that connect pairs of joints in the human skeleton, denoted by $\mathcal{G}=\{\bm{\theta}=(\theta_1,\dots,\theta_L) \: | \: \theta_l \in \mathbb{R}^2,\Vert\theta_l\Vert_2=1 \: \forall l \in [L]\}$. We assume that plausible 2D human poses lie on a manifold embedded in this pose space $\mathcal{G}$. We use a function $g:\mathcal{G} \xrightarrow{} \mathbb{R}^+$, which maps a pose to a non-negative scalar, to represent the manifold of plausible poses as the zero-level set:
\begin{equation}
    \mathcal{P} = \{\bm{\theta} \in \mathcal{G} \: | \: g(\bm{\theta})=0\},
\end{equation}
where $g$ represents the unsigned distance to the manifold. We construct this distance function by first encoding the pose using a hierarchical network $g_\textrm{enc}$ that encodes the human pose based on its anatomical structure~\cite{mihajlovic2021leap}, and subsequently, use $g_\textrm{dec}$ to predict the distance based on the pose representation.

Specifically, for a given pose $\bm{\theta}$, we encode it as follows,
\begin{equation}
    \mathbf{v}_1 = g_\textrm{enc}^1(\theta_1), \quad \: \mathbf{v}_l = g_\textrm{enc}^l(\theta_l,\mathbf{v}_{\Omega(l)}), l \in \{2,\dots,L\}. 
\end{equation}
Here, $\Omega(l)$ is a function that maps the index of each orientation vector to its parent orientation vector in the kinematic chain of the human skeleton. We obtain the overall pose encoding as $\mathbf{p} = [\mathbf{v}_1|\dots|\mathbf{v}_L]$ by concatenating all the individual orientation encodings. This pose encoding is then processed by $g_\textrm{dec}:\mathbb{R}^{d.L}\xrightarrow{}\mathbb{R}^+$, which predicts the unsigned distance for the given pose representation $\mathbf{p}$. A lower distance value for a pose implies that the configuration of joints is more likely to be a plausible human pose.

\subsubsection{Pose Prior Training}
To train the parametric prior $g$, we rely on an auxiliary dataset of $M$ human poses $\mathcal{D}_\textrm{A}=\{\bm{\theta}^i\}_{i=1}^{N_\textrm{A}}$, where $\bm{\theta}^i=(\theta^i_1,\dots,\theta^i_L)$. Importantly, these poses are not associated with their corresponding RGB images, which preserves the privacy aspect of the method. Additionally, storing only the pose coordinates instead of entire images makes data storage much more efficient and feasible.

We adopt a supervised approach to train $g$ to predict the $L2$ distance to the plausible pose manifold for a given pose. To achieve this, we construct a dataset $\Tilde{\mathcal{D}}=\{(\bm{\theta}_i,d_i)\}_{i=1}^M$, consisting of pose and distance pairs, from $\mathcal{D}_\textrm{A}$. As the poses from $\mathcal{D}_\textrm{A}$ lie on the desired manifold, we assign $d=0$ to all poses in the dataset. To diversify our training samples, we randomly generate negative samples with distance $d>0$ by perturbing the poses from $\mathcal{D}_\textrm{A}$ with noise. 

We train the network with the standard $L1$ loss,
\begin{equation}
    \mathcal{L}_\textrm{dist} = \sum_{(\bm{\theta},d)\in\Tilde{D}}\Vert g(\bm{\theta})-d \Vert_1.
\end{equation}

More details on the training process of the prior are provided in Section~\ref{sec:protocols}.

\subsubsection{Adaptation using Pose Prior}
We leverage the trained pose prior $g$ to regularize the adaptation process by incentivizing the pose estimator to generate pseudo-labels that resemble plausible human poses. 

Given the heatmaps $\{h^t_k\}_{k=1}^K$ generated for each keypoint by the student model $f^t_\textrm{stu}$ for a target image $\bm{x}$, we calculate the corresponding orientation vectors in a differentiable manner to evaluate the plausibility of the predicted pose using the prior. First, we renormalize each heatmap to a probability distribution via spatial softmax and condense it to a point by computing the spatial expected value of the latter. For computational efficiency, we carry this out in a separable manner along the two spatial dimensions. Namely, assuming $u=(u_1, u_2)$ to be the two components of each pixel coordinate, we set
\begin{equation}
    u_{i}^k = \frac{\sum_{u_i}u_i e^{h^t_k(u_i)}}{\sum_{u_i}e^{h^t_k(u_i)}}, \quad h^t_k(u_i) = \sum_{u_j}h^t_k(u_1,u_2),
\end{equation}
where $i=1,2$ and $j=2,1$ respectively. Next, we use the pose coordinates to determine the orientation vectors between pairs of connected keypoints. Specifically, for every pair $(a,b)$ of connected keypoints in the human skeleton (denoted by the set $\mathcal{E}$), we calculate the unit vector $\theta$ in the direction from $u^a$ to $u^b$, where $u^a$ and $u^b$ are the estimated softmax coordinates of keypoints $a$ and $b$, respectively, i.e.,
\begin{equation}
    \theta = \frac{u^a - u^b}{\Vert u^a - u^b\Vert_2},  \quad \forall (a,b) \in \mathcal{E}.
\end{equation}
Finally, we use the prior as a regularization term to minimize the distance of the current pose from our learned manifold,
\begin{equation}
    \mathcal{L}_\textrm{prior} = \frac{1}{|\mathcal{B}|}\sum_{\bm{x}\in\mathcal{B}}g(T(f^t_\textrm{stu}(\bm{x}))),
\end{equation}
where $T(\cdot)$ converts the predicted heatmaps to the orientation vector format required as input by the prior.

\subsection{Overall Adaptation}
The final training objective for the student model $f_\textrm{stu}$ can be expressed as:
\begin{equation}
    \min_{f_\textrm{stu}} \: \mathcal{L}_\textrm{out} + \lambda_1\mathcal{L}_\textrm{feat} + \lambda_2\mathcal{L}_\textrm{prior}.
\end{equation}
Here, $\lambda_1$ and $\lambda_2$ are hyper-parameters that control the influence of feature space adaptation and prior regularization, respectively. The teacher model $f_\textrm{tea}$ is updated asynchronously by computing an exponential moving average of the student model weights as shown in Equation~\ref{eq:ema}.

%% file: sections/4_experiments.tex
\section{Experiments}
In this section, we demonstrate \texttt{POST}'s ability to adapt a 2D human pose estimation model to a target domain using only unlabeled data from that domain. We conduct experiments on three domain adaptive scenarios and compare with state-of-the-art domain adaptation baselines that utilize source data during adaptation. We also conduct extensive analysis to analyze the contribution and interaction between each component in our framework.

\subsection{Datasets}
\noindent \textbf{SURREAL:} SURREAL~\cite{varol17_surreal} is a large-scale dataset of synthetically generated images of people rendered from 3D sequences of human motion capture data against indoor backgrounds. It contains over 6 million frames, making it one of the largest and most diverse datasets of its kind.
\vskip 4pt
\noindent \textbf{Human3.6M:} Human3.6M~\cite{ionescu2013human3} is a real-world video dataset captured in indoor environments, comprising 3.6 million frames. The dataset features human subjects performing various actions. In order to reduce redundancy and computational complexity, we down-sampled the videos from 50fps to 10fps as per the approach proposed in~\cite{jiang2021regressive}. For training, we follow the standard protocol proposed in~\cite{kim2022unified} and use 5 subjects (S1, S5, S6, S7, S8), while the remaining 2 subjects (S9, S11) are reserved for testing. 
\vskip 4pt
\noindent \textbf{LSP:} Leeds Sports Pose (LSP)~\cite{johnson2010clustered} is a real-world dataset that contains 2,000 images with annotated human body joint locations collected from sports activities. The images in LSP are captured in the wild, featuring a wide variety of human poses that are often challenging to detect.
\vskip 4pt
\noindent \textbf{BRIAR:} BRIAR~\cite{Cornett_2023_WACV} is a cutting-edge biometric dataset featuring a large-scale collection of videos of human subjects captured in extremely challenging conditions. The videos are recorded at varying distances, \emph{i.e.}, close range, 100m, 200m, 400m, 500m, and unmanned aerial vehicles (UAV), with each video lasting around 90 seconds. We randomly sample 20 frames from each sequence for each of the 158 subjects for our experiments. 

\subsection{Experiment Protocols} \label{sec:protocols}
\noindent \textbf{Pose estimation model.} We adopt the Simple Baseline~\cite{xiao2018simple} as our pose estimation model, with the ResNet-101~\cite{he2016deep} as the backbone. To train the model, we use the Adam optimizer~\cite{kingma2014adam} with a base learning rate of $1e-4$, scheduled to decrease to $1e-5$ after 5 epochs and to $1e-6$ after 20 epochs. The model is trained for 30 epochs. We use a batch size of 32 and run 500 iterations per epoch. To threshold the model predictions, we set the confidence thresholding ratio $p$ to 0.5. To augment the images during training, we follow~\cite{kim2022unified} and use rotation, translation, shear, and Gaussian blur. The hyper-parameters $\lambda_1$ and $\lambda_2$ are set to $1e-3$ and $1e-6$, respectively.
\vskip 4pt
\noindent \textbf{Pose prior model.}
The training of the parametric prior follows a multi-stage approach that involves using different types of training samples. Initially, we use a combination of manifold poses $\bm{\theta_m}$ and non-manifold poses $\bm{\theta_{nm}}$ with a considerable distance from the desired manifold. Over the course of training, we gradually increase the number of non-manifold poses $\bm{\theta_{nm}}$ with a small distance from the manifold. This enables our model to first learn a smooth surface and then gradually incorporate finer details as training progresses. We create these non-manifold poses $\bm{\theta_{nm}}$ by injecting noise into the manifold poses $\bm{\theta_{m}}$ obtained from the auxiliary dataset. Specifically, we sample directional noise from the Von-Mises distribution \cite{gatto2007generalized} and add it to the manifold poses in order to obtain the implausible poses.

The architecture for the encoder $g_\textrm{enc}$ consists of a 2-layer MLP with an output feature size of $d=6$ for each orientation vector, similar to~\cite{mihajlovic2021leap}. The distance field network $g_\textrm{dec}$ is implemented as a 5-layer MLP. Given its large size and diverse poses, we train the prior for our primary experiments using the SURREAL dataset. 

\subsection{Comparison with Baselines}
\noindent \textbf{Baselines.} We evaluate the performance of our proposed method against several state-of-the-art domain adaptive frameworks. This includes adversarial learning-based feature alignment methods, such as \textit{DAN}\cite{long2015learning}, \textit{DD}\cite{zhang2019bridging}, and \textit{RegDA}\cite{jiang2021regressive}. Additionally, we consider approaches based on pseudo-labeling, namely \textit{CCSSL}\cite{mu2020learning}, and \textit{UDAPE}~\cite{kim2022unified}. It is worth noting that all these methods employ the source data during the adaptation process. To establish a comprehensive performance baseline, we report the results of two additional baselines: \textit{Oracle} and \textit{Source only}. The \textit{Oracle} baseline represents the upper bound of the model's performance, achieved by training the model jointly with target 2D annotations. On the other hand, \textit{Source only} represents the model's performance when it is directly applied to the target domain without any adaptation. 

\noindent \textbf{Metrics.} We adopt the evaluation metric of Percentage of Correct Keypoint (PCK) for all experiments and report PCK@0.05 that measures the ratio of correct prediction within a range of 5\% with respect to the image size.

\input{tables/lsp}
\input{tables/h36m}
\subsection{Results}
\begin{figure}[t]
    \centering
    \includegraphics[width=0.7\columnwidth]{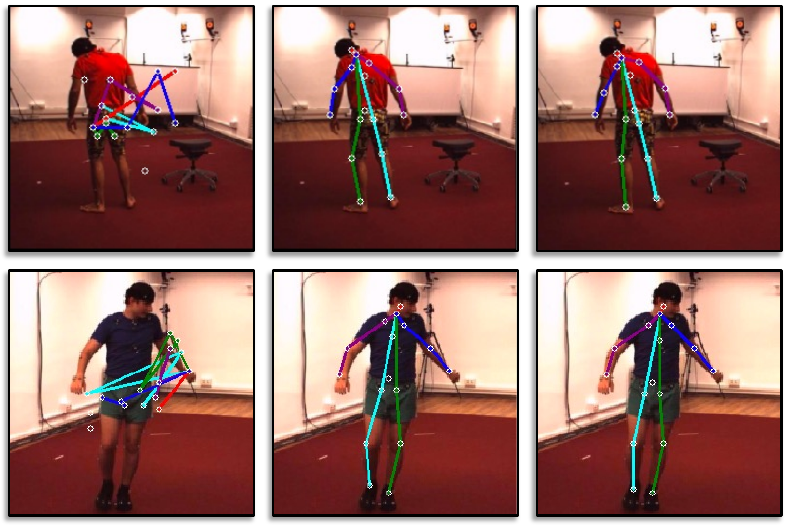}
    \caption{\textbf{Qualitative results on SURREAL $\rightarrow$ Human3.6M.} We demonstrate sample results on the Human3.6M dataset. For each row, the leftmost shows the \textit{Source only} prediction, the middle one shows the \textit{UDAPE}~\cite{kim2022unified} prediction, and the rightmost shows the prediction made \texttt{POST}.}
    \label{fig:h36m}
\end{figure}

\begin{figure}[t]
    \centering
    \includegraphics[width=0.6\columnwidth]{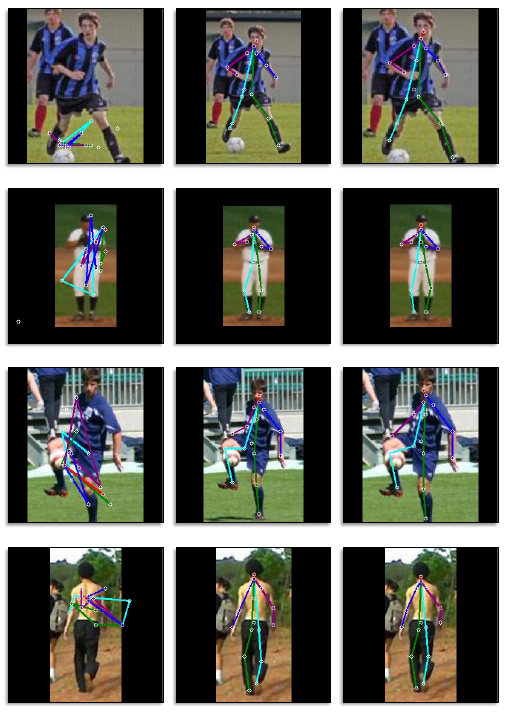}
    \caption{\textbf{Qualitative results on SURREAL $\rightarrow$ LSP.} We demonstrate sample results on the LSP dataset. For each row, the leftmost shows the \textit{Source only} prediction, the middle one shows the \textit{UDAPE}~\cite{kim2022unified} prediction, and the rightmost shows the prediction made \texttt{POST}.}
    \label{fig:lsp}
\end{figure}

\begin{figure*}[t]
    \centering
    \includegraphics[width=0.8\textwidth]{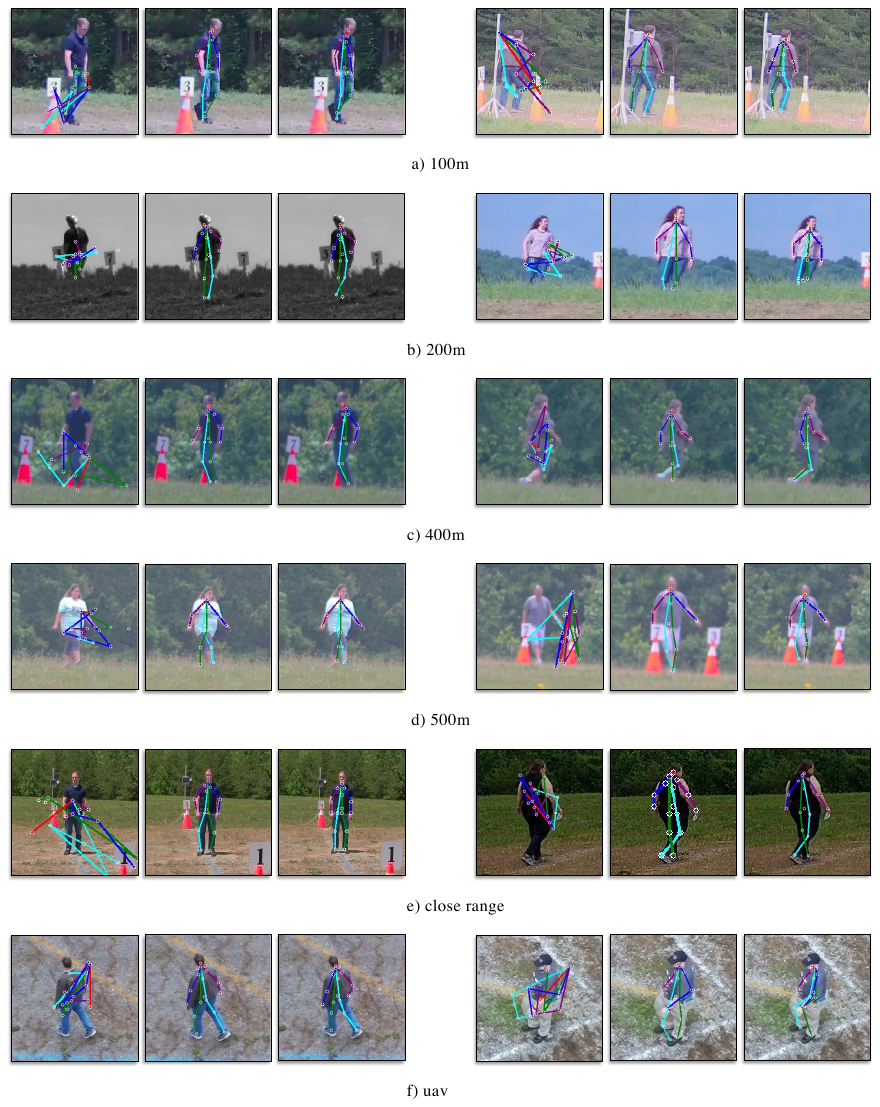}
    \caption{\textbf{Qualitative results on SURREAL $\rightarrow$ BRIAR.} We demonstrate sample results on BRIAR at all ranges. For each range, we display three images: the leftmost shows the \textit{Source only} prediction, the middle one shows the \textit{UDAPE}~\cite{kim2022unified} prediction, and the rightmost shows the prediction using \texttt{POST}. Although \texttt{POST} does not use source data for adaptation, it is able to match the predictions produced by UDAPE, which uses source data.}
    \label{fig:briar}
\end{figure*}

\noindent \textbf{Quantitative results.} We evaluate \texttt{POST} in two adaptation scenarios: SURREAL$\rightarrow$LSP and SURREAL$\rightarrow$Human3.6M, and report the quantitative results in Table~\ref{tab:lsp} and Table~\ref{tab:h36m}, respectively. Specifically, we report the PCK@0.05 on 16 keypoints of the human body, including shoulders (sld.), elbows (elb.), wrists, hips, knees, and ankles. Our method achieves comparable results to many recent approaches that leverage source data for adaptation. Among these methods, UDAPE~\cite{kim2022unified} achieves the highest performance on both cases, with our framework achieving a close second and only 2 percentage points behind on average. Notably, we outperform every other method, including the recently proposed RegDA~\cite{jiang2021regressive} approach, by a significant margin of up to 5.7 percentage points. It is worth noting that not only do these methods require source data, but also involve additional models such as discriminators or style transfer modules, and unstable adversarial training. In contrast, our framework is lightweight, only involves pseudo-label training, and utilizes a simple prior model that can be trained offline. 
\vskip 4pt
\noindent \textbf{Qualitative results.} In addition to quantitative results, we also present qualitative results on SURREAL$\rightarrow$Human3.6M in Figure~\ref{fig:h36m} and on SURREAL$\rightarrow$LSP in Figure~\ref{fig:lsp}. Also, we present only visual results on the SURREAL$\rightarrow$BRIAR adaptation scenario since pose annotations are absent in the BRIAR dataset. Figure~\ref{fig:briar} displays the predicted human poses on images taken from six different imaging ranges in the BRIAR dataset. While using the source model directly produces completely inaccurate poses, \texttt{POST} can accurately localize the keypoint locations, even in the presence of occlusions and atmospheric turbulence. Our approach can also accurately reconstruct poses even when the human is imaged from an elevated perspective, resulting in a high camera angle (Figure~\ref{fig:briar}\textcolor{red}{.f}). Notably, our approach achieves this without using any source data for adaptation. Our results are comparable to those produced by UDAPE~\cite{kim2022unified}, which utilizes source data for adaptation. Additional qualitative results can be found in the appendix.

\input{tables/ablation_prior}
\subsection{Ablation Studies}
\noindent \textbf{Effect of auxiliary dataset.} We conduct experiments to evaluate the impact of the choice of the auxiliary dataset (used to train the prior) on downstream adaptation. The results are presented in Table~\ref{tab:ab_prior}. Our findings indicate that \texttt{POST} is robust to the choice of the auxiliary dataset, with performance differences of $\sim 0.5$ percentage points.
\vskip 4pt
\noindent \textbf{Effect of loss terms.} We conduct an experiment to evaluate the performance of each component of our framework. The results on the SURREAL$\rightarrow$Human3.6M adaptation scenario are presented in Table~\ref{tab:ab_loss_h36m}. Our findings indicate that in addition to prediction space adaptation, feature space adaptation also plays a crucial role in enabling effective unsupervised learning from pseudo-labels. Moreover, we observed that the human pose prior brings additional improvements, thus validating our hypothesis that noisy pose pseudo-labels can be refined implicitly by a prior in the absence of source data. Overall, our results demonstrate the effectiveness of our framework and the importance of each of its components in achieving state-of-the-art performance. 
\vskip 4pt
\noindent \textbf{Cross-dataset performance of prior.} We evaluate the ability of the learned prior to handle distribution shift separately from the adaptation performance. Specifically, we assess the robustness of the prior by computing pose scores across datasets, and the results are presented in Figure~\ref{fig:prior_transfer}. The plot demonstrates that our prior is effective in scoring plausible (real) poses with low scores and scoring implausible poses with higher scores. Note that the bell curve shape of the noisy pose scores is due to the Von-Mises noise added to create the noisy poses. 
\vskip 4pt
\noindent \textbf{Effect of thresholding.} We analyze the impact of the pseudo-label threshold $\tau$ on the adaptation performance in Table~\ref{tab:ab_tau}. The results on SURREAL$\rightarrow$Human3.6M reveal that as we increase this ratio, the performance gradually decreases. This can be attributed to the fact that higher thresholding ratios tend to include lower confident predictions as pseudo-labels, which can negatively impact adaptation.

\input{tables/ablation_loss}
\input{tables/ablation_tau}

%% file: tables/lsp.tex
\begin{table}[]
\centering
\caption{\textbf{PCK@0.05 on SURREAL $\rightarrow$ LSP.} (Best value is in
\textcolor{red}{red} color, while the second best value is in \textcolor{blue}{blue} color.)}
\label{tab:lsp}
\vskip 0.08in
\resizebox{\columnwidth}{!}{%
\begin{tabular}{lccccccca}
\toprule
\textbf{Method} & \textbf{SF} & \textbf{Sld.} & \textbf{Elb.} & \textbf{Wrist} & \textbf{Hip} & \textbf{Knee} & \textbf{Ankle} & \textbf{Avg.} \\ \midrule
Source only
& -
& 51.5     
& 65.0         
& 62.9  
& 68.0        
& 68.7       
& 67.4       
& 63.9  \\
Oracle
& -
& -    
& -
& -
& -
& -
& -
& -  \\\midrule
DAN
& \color{red}\xmark
& 52.2     
& 62.9 
& 58.9 
& 71.0 
& 68.1 
& 65.1 
& 63.0  \\
DD
& \color{red}\xmark
& 28.4     
& 65.9 
& 56.8 
& 75.0 
& 74.3 
& 73.9 
& 62.4  \\
RegDA
& \color{red}\xmark
& 62.7     
& 76.7 
& 71.1 
& 81.0 
& 80.3 
& 75.3 
& 74.6  \\
CCSSL
& \color{red}\xmark
& 36.8 
& 66.3 
& 63.9 
& 59.6 
& 67.3 
& 70.4 
& 60.7  \\
UDAPE
& \color{red}\xmark
& \textcolor{Red}{69.2} 
& \textcolor{Red}{84.9} 
& \textcolor{Red}{83.3}  
& \textcolor{Red}{85.5} 
& \textcolor{Red}{84.7} 
& \textcolor{Red}{84.3}  
& \textcolor{Red}{82.0}      \\ \midrule
POST
& \color{ForestGreen}\cmark
& \textcolor{Blue}{66.5}    
& \textcolor{Blue}{83.9}
& \textcolor{Blue}{81.0}
& \textcolor{Blue}{84.6}
& \textcolor{Blue}{83.1}
& \textcolor{Blue}{82.6}
& \textcolor{Blue}{80.3}  \\\bottomrule
\end{tabular}%
}
\end{table}

%% file: tables/h36m.tex
\begin{table}[]
\centering
\caption{\textbf{PCK@0.05 on SURREAL $\rightarrow$ Human3.6M.}}
\label{tab:h36m}
\vskip 0.08in
\resizebox{\columnwidth}{!}{%
\begin{tabular}{lccccccca}
\toprule
\textbf{Method} & \textbf{SF} & \textbf{Sld.} & \textbf{Elb.} & \textbf{Wrist} & \textbf{Hip} & \textbf{Knee} & \textbf{Ankle} & \textbf{Avg.} \\ \midrule
Source only
& -
& 69.4 
& 75.4     
& 66.4       
& 37.9       
& 77.3      
& 77.7        
& 67.3  \\
Oracle
& -
& 95.3 
& 91.8 
& 86.9 
& 95.6 
& 94.1 
& 93.6  
& 92.9   \\\midrule
DAN
& \color{red}\xmark
& 68.1 
& 77.5 
& 62.3 
& 30.4 
& 78.4   
& 79.4   
& 66.0  \\
DD
& \color{red}\xmark
& 71.6  
& 83.3 
& 75.1 
& 42.1 
& 76.2 
& 76.1 
& 70.7     \\
RegDA
& \color{red}\xmark
& 73.3 
& 86.4 
& 72.8    
& \textcolor{Red}{54.8} 
& 82.0 
& \textcolor{Blue}{84.4}  
& 75.6  \\
CCSSL
& \color{red}\xmark
& 44.3      
& 68.5 
& 55.2 
& 22.2 
& 62.3 
& 57.8 
& 51.7  \\
UDAPE
& \color{red}\xmark
& \textcolor{Blue}{78.1} 
& \textcolor{Red}{89.6}
& \textcolor{Red}{81.1} 
& \textcolor{Blue}{52.6} 
& \textcolor{Red}{85.3} 
& \textcolor{Red}{87.1} 
& \textcolor{Red}{79.0}   \\ \midrule
POST
& \color{ForestGreen}\cmark
& \textcolor{Red}{81.3}    
& \textcolor{Blue}{88.5}
& \textcolor{Blue}{77.4}
& 46.1
& \textcolor{Blue}{83.4}
& 83.4
& \textcolor{Blue}{76.7}  \\\bottomrule
\end{tabular}%
}
\end{table}

%% file: tables/ablation_prior.tex
\begin{table}[]
\centering
\caption{\textbf{Effect of auxiliary dataset.} We evaluate the effect of auxiliary dataset on downstream adaptation tasks.}
\vskip 0.08in
\resizebox{\columnwidth}{!}{%
\begin{tabular}{lcccccca}
\toprule
\multicolumn{8}{c}{\textsc{SURREAL}$\rightarrow$\textsc{Human3.6M}} \\[2pt]
\textbf{Aux. dataset} & \textbf{Sld.} & \textbf{Elb.} & \textbf{Wrist} & \textbf{Hip} & \textbf{Knee} & \textbf{Ankle} & \textbf{Avg.} \\ 
\midrule
SURREAL
& 81.3   
& 88.5
& 77.4
& 46.1
& 83.4
& 83.4
& 76.7 \\
Human3.6M
& 80.9
& 88.0
& 77.2
& 45.0
& 83.1
& 82.8
& 76.2 \\
\midrule
\multicolumn{8}{c}{\textsc{SURREAL}$\rightarrow$\textsc{LSP}} \\[2pt]
\textbf{Aux. dataset} & \textbf{Sld.} & \textbf{Elb.} & \textbf{Wrist} & \textbf{Hip} & \textbf{Knee} & \textbf{Ankle} & \textbf{Avg.} \\ 
\midrule
SURREAL
& 66.5
& 83.9
& 81.0
& 84.6
& 83.1
& 82.6
& 80.3  \\
Human3.6M
& 66.1
& 83.6
& 80.7
& 84.4
& 83.1
& 82.5
& 80.1 \\
\bottomrule
\end{tabular}%
}
\label{tab:ab_prior}
\end{table}

%% file: tables/ablation_loss.tex
\begin{table}[]
\centering
\caption{\textbf{Effect of each loss term.} We evaluate the contribution of each loss term on SURREAL$\rightarrow$Human3.6M. }
\vskip 0.08in
\resizebox{\columnwidth}{!}{%
\begin{tabular}{ccccccccca}
\toprule
$\mathcal{L}_\textrm{out}$ & $\mathcal{L}_\textrm{feat}$ & $\mathcal{L}_\textrm{prior}$  & \textbf{Sld.} & \textbf{Elb.} & \textbf{Wrist} & \textbf{Hip} & \textbf{Knee} & \textbf{Ankle} & \textbf{Avg.} \\ 
\midrule
\xmark
& \xmark
& \xmark
& 69.4 
& 75.4     
& 66.4       
& 37.9       
& 77.3      
& 77.7        
& 67.3 \\
\cmark 
& \xmark
& \xmark
& 77.9 
& 86.7
& 73.7
& 38.8
& 83.0
& 84.3
& 74.1\\
\cmark 
& \cmark 
& \xmark
& \textcolor{red}{81.7}  
& \textcolor{blue}{87.1}
& \textcolor{blue}{75.2}
& \textcolor{blue}{44.3}
& \textcolor{blue}{82.3}
& \textcolor{blue}{82.2}
& \textcolor{blue}{75.5}\\
\cmark 
& \cmark 
& \cmark 
& \textcolor{blue}{81.3}   
& \textcolor{red}{88.5}
& \textcolor{red}{77.4}
& \textcolor{red}{46.1}
& \textcolor{red}{83.4}
& \textcolor{red}{83.4}
& \textcolor{red}{76.7} \\
\bottomrule
\end{tabular}%
}
\label{tab:ab_loss_h36m}
\end{table}

%% file: tables/ablation_tau.tex
\begin{table}[]
\centering
\scriptsize
\caption{\textbf{Effect of $\bm{\tau}$.} We evaluate adaptation performance on SURREAL$\rightarrow$Human3.6M as $\tau$ is varied. }
\vskip 0.08in
\begin{tabular}{ccccc}
\toprule
$\tau=0.1$ & $\tau=0.3$ & $\tau=0.5$  & $\tau=0.7$ & $\tau=0.9$ \\ 
\midrule
76.4
& 76.5
& 76.7
& 75.3
& 73.8 \\
\bottomrule
\end{tabular}%
\label{tab:ab_tau}
\end{table}

%% file: sections/5_conclusion.tex
\section{Conclusion}
We address the problem of adapting a pre-trained 2D human pose estimator to a new target domain with unlabeled target data. To this end, we propose a self-training algorithm, \texttt{POST}, that leverages a Mean Teacher framework to enforce both prediction-level and feature-level consistency between a pair of student and teacher models. Our approach incorporates a human pose prior that captures the manifold of possible poses in a high-dimensional space, which helps to refine noisy pseudo-labels generated during self-training. We evaluate our method on three challenging adaptation scenarios and show that it achieves competitive performance compared to existing UDA methods that have access to source data.
\vskip 6pt
\noindent \textbf{Acknowledgements.} This research is based upon work supported in part by the Office of the Director of National Intelligence (ODNI), Intelligence Advanced Research Projects Activity (IARPA), via [2022-21102100007]. The views and conclusions contained herein are those of the authors and should not be interpreted as necessarily representing the official policies, either expressed or implied, of ODNI, IARPA, or the U.S. Government. The U.S. Government is authorized to reproduce and distribute reprints for governmental purposes notwithstanding any copyright annotation therein.

\begin{figure}[t]
    \centering
    \includegraphics[width=0.8\columnwidth]{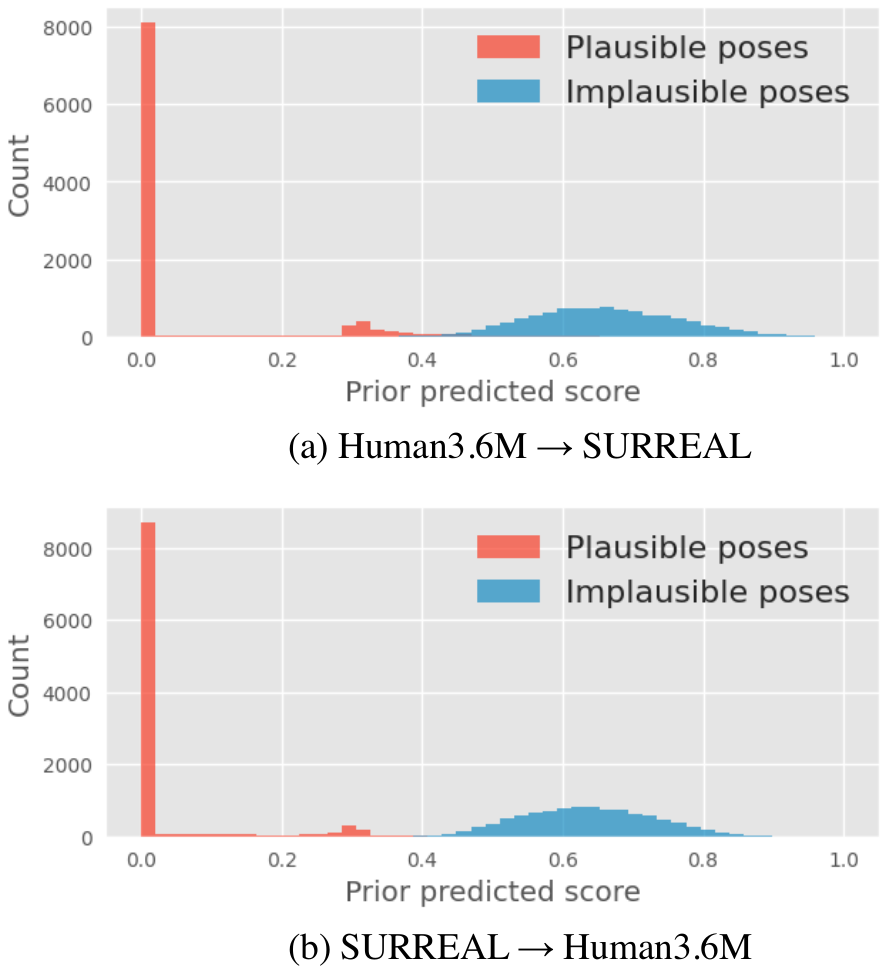}
    \caption{\textbf{Cross-dataset prior transfer.} We plot the histogram of scores predicted by the prior. The prior can clearly demarcate plausible poses from implausible poses across datasets.}
    \label{fig:prior_transfer}
\end{figure}

%% file: sections/supp.tex
\section{Processing auxiliary dataset}
In the experiment section of the main paper, we briefly described how to process the auxiliary dataset $\mathcal{D}_\textrm{A}$ to train the prior. Specifically, we generated a set of both plausible and implausible poses and calculated the corresponding distance values from the pose manifold $\Tilde{\mathcal{D}}=\{(\bm{\theta}_i,d_i)\}_{i=1}^M$. 

For poses derived from the auxiliary dataset, we label them as plausible poses and assign a distance value of $d=0$. To generate an implausible pose $\bm{\theta_{nm}}$, we first randomly sample a pose $\bm{\theta_m}\sim \mathcal{D}_\textrm{A}$ and convert the sampled pose to polar coordinates, 
\begin{align*}
    u_1 = \arccos(\bm{\theta_m}^1) \\
    u_2 = \arcsin(\bm{\theta_m}^2).   
\end{align*}
 Next, we sample noise from the Von-Mises distribution, 
\begin{equation*}
    n_1, n_2 \sim f(n|\mu,\kappa) \quad \textrm{where } f(n|\mu,\kappa) = \frac{\exp{(\kappa\cos(n-\mu))}}{2\pi I_0(\kappa)},
\end{equation*}
and add it to the coordinates to obtain the new pose $\bm{\theta_{nm}}$,
\begin{align*}
    \bm{\theta_{nm}}^1 &= \cos(u_1 + n_1)\\
    \bm{\theta_{nm}}^2 &= \sin(u_2 + n_2).   
\end{align*}
In  our experiments we set $\mu=0$ and sample $\kappa$ randomly from the set $\{2,4,8\}$.

We employ the nearest neighbor strategy described in~\cite{tiwari22posendf} to assign a distance value to each synthetically generated pose. To accomplish this, we first use FAISS~\cite{johnson2019billion} and L2 distances to approximate the $k'$ nearest neighbors of the pose from the set of clean poses. From these neighbors, we identify the exact $k$ nearest ones. In our approach, we set $k'=500$ and $k=5$. Finally, we determine the ground truth distance by calculating the average of the $k$ smallest distances.

\begin{figure}[t]
    \centering
    \includegraphics[width=0.7\columnwidth]{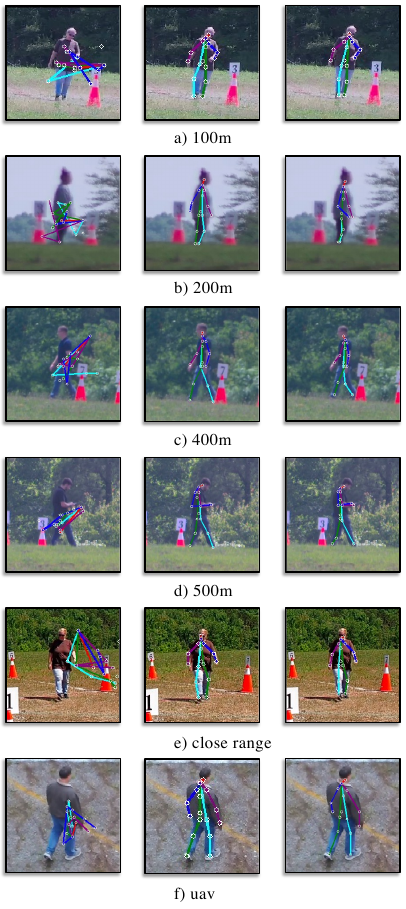}
    \caption{\textbf{Qualitative results on SURREAL $\rightarrow$ BRIAR.} We demonstrate sample results on the BRIAR dataset at all ranges. For each range, we display three images: the leftmost shows the \textit{Source only} prediction, the middle one shows the \textit{UDAPE}~\cite{kim2022unified} prediction, and the rightmost shows the prediction made by our framework.}
    \label{fig:briar_supp}
\end{figure}
\section{Qualitative results}
We present additional visual results on the SURREAL$\rightarrow$BRIAR scenario in Figure~\ref{fig:briar_supp}. Although our approach does not use any source data for adaptation, it is able to match the predictions produced by UDAPE~\cite{kim2022unified}, which uses source data.